\documentclass[letterpaper]{article} 
\usepackage{aaai25}  
\usepackage{times}  
\usepackage{helvet}  
\usepackage{courier}  
\usepackage[hyphens]{url}  
\usepackage{graphicx} 
\usepackage{natbib}  
\usepackage{tcolorbox}

\usepackage{caption} 
\usepackage{algorithm}
\usepackage{algorithmic}
\usepackage{amsmath}
\usepackage{newfloat}
\usepackage{listings}
\usepackage{tikz}
\def\checkmark{\tikz\fill[scale=0.4](0,.35) -- (.25,0) -- (1,.7) -- (.25,.15) -- cycle;} 

\DeclareCaptionStyle{ruled}{labelfont=normalfont,labelsep=colon,strut=off} 
\floatstyle{ruled}
\newfloat{listing}{tb}{lst}{}
\floatname{listing}{Listing}
\title{A Novel Approach to Balance Convenience and Nutrition in Meals With Long-Term Group Recommendations and Reasoning on Multimodal Recipes and its Implementation in  BEACON}
\author{
    Vansh Nagpal,
    Siva Likitha Valluru,
    Kausik Lakkaraju, 
    Nitin Gupta, \\
    Zach Abdulrahman\equalcontrib, 
    Andrew Davison\equalcontrib, 
    Biplav Srivastava
}
\affiliations{

    \{vnagpal@email., svalluru@email., kausik@email., niting@email.,zja@email., ajd8@email.,  biplav.s@\}sc.edu \\
    University of South Carolina     
}

\usepackage{color}
\usepackage{booktabs}
\usepackage{csvsimple}

\newcommand{\vansh}[1]{{\color{magenta}~{\em Comment by Vansh: #1}}}

\usepackage{xcolor}

\lstdefinelanguage{JSON}{
    string=[s]{"}{"},
    stringstyle=\color{black},
    numbers=left,
    numberstyle=\small,
    basicstyle=\ttfamily,
    keywordstyle=\color{blue},
    keywords={false,true,null},
    comment=[l]{:},
    commentstyle=\color{black}
}

\begin{document}

\maketitle

\begin{abstract}
A common, yet regular, decision made by people, whether healthy or with any health condition, is to decide what to have in meals like breakfast, lunch, and dinner, consisting of a combination of foods for appetizer, main course, side dishes, desserts, and beverages. However, often this decision is seen as a trade-off between nutritious choices (e.g.,  salt and sugar levels, nutrition content) or convenience (e.g., cost and speed to access, cuisine type,  food source type). We present a data-driven solution for meal recommendations that considers customizable meal configurations and time horizons. This solution balances user preferences while taking into account a food’s constituents and cooking process. Beyond the problem formulation, our contributions include introducing goodness measures, a recipe conversion method from text to the recently introduced multimodal rich recipe representation (R3) format,  learning methods using contextual bandits that show promising preliminary results, and the prototype, usage-inspired, BEACON system.
\end{abstract}

%

\section{Introduction} \label{sec:introduction}
Although it is well known that nutritious foods are essential to a person's health, the actual adherence to dietary requirements is quite poor across the world. In fact, according to a recent meta-survey \cite{leme2021adherence}, almost 40\% of the population across high-, low-, and medium-income countries do not adhere to their national food-based dietary guidelines, often prioritizing convenience over nutrition needs. Previous studies have shown that adhering to a provided meal plan instead of a self-selected one reduces the risk for adverse health conditions  \cite{metz1997dietary}.  
Some people prefer getting food recommendations from their friends or family, and others turn to online recommender systems \cite{yang2017yum} or even Large Language Models (LLMs) \cite{rostami2024integrated} as they have become easily available in the form of chatbots, albeit being inaccurate, misleading, or not wholly informed.  For example, authors in \cite{papastratis2024can} found that ChatGPT alone is not a reliable tool for meal recommendation when assessing ChatGPT-based recommenders for balanced diets in non-communicable disease (NCD) patients. 

\begin{figure*}
    \centering
    \includegraphics[width=0.99\textwidth]{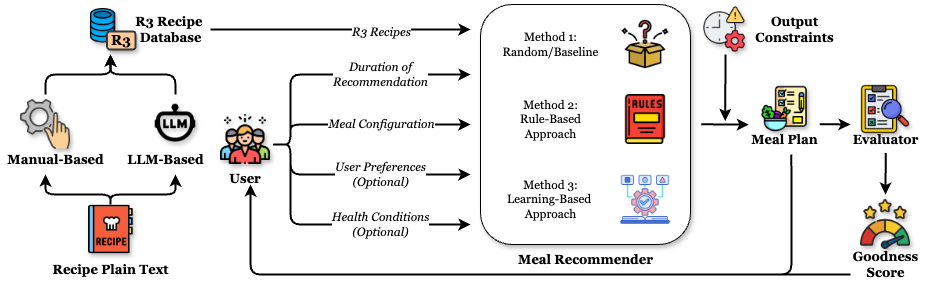}
    \caption{A brief depiction of solution components used in the BEACON prototype.}
    \label{fig:archi}
\end{figure*}

We seek to help the general population decide on meal choices while nudging them towards healthy choices by leveraging data from online recipes, domain knowledge about meals and how they are configured from foods, and user preferences (Figure~\ref{fig:archi}).
In doing so, we recognize the reality that people want to explore a variety of foods, and a {\em long-horizon meal recommender} can act as a trusted companion seeking to keep the user well-informed even when they deviate from nutrition guidelines. 




%

Our contributions are that we: (1)~introduce a novel approach to the meal recommendation problem, accounting for variable meal configurations and flexible time horizons, and propose innovative quantitative metrics to evaluate the framework and benchmark its performance against relevant baselines, (2)~adopt the multi-modal R3 framework \cite{pallagani2022rich} to convert recipes from two prominent fast food chains into R3 representations, leveraging various LLM-based methodologies to ensure robust and accurate transformation, and (3)~present the design of the BEACON meal recommender system, showcasing its potential through a compelling and practical use case to effectively balance both convenience and nutrition.


In the remainder of the paper, we provide relevant background in automated recommendations of personalized meals and then discuss our problem formulation, key solution components including data (recipe representation and format conversion) and meal recommendation, and their evaluation. We then describe a prototype implementation of the solution in the BEACON system, along with the supported use cases, and conclude with a discussion of practical considerations and avenues for future extensions.

\section{Related Work} \label{sec:related_work}

There is a large body of literature on recommendation methods for single items \cite{su2009survey,cremonesi2010performance}. In many practical situations,  a \textit{group} of items has to be recommended, as in the case of team formation. Here, although the group problem can be treated as a special case of \textit{sequential} single-item recommendation problem \cite{srivastava2022ultra}, better results are found when treating them as a group \cite{valluru2024promoting, valluru2024ultra, valluru2024ai}. Our work falls in the latter category.


There are food recommendation systems in literature that seek to 
guide users based on their dietary preferences, health conditions, and nutritional requirements. 
They offer tailored guidance on nutritionally balanced meal options, ensuring that users consume the right combination of macronutrients and micronutrients to support their physical activity and overall health~\cite{bekdash2024epigenetics}. 
Examples include \textit{single} food items~\cite{yang2017yum, ge2015using}.
They may also help with weight management, such as portion control and unhealthy food cravings~\cite{dunn2018mindfulness}, by offering satisfying yet calorie-conscious alternatives. 

Considering the problem of meal plan recommendation, the representation of the food items is a crucial portion of the overall system. Many works utilize text descriptions \cite{li2020reciptor} of food items or recipes, which limits machine readability. Alternatively, we use a structured form of recipes in R3 format \cite{pallagani2022rich} which covers the content (food description) and also the process of preparing it, spanning text and image. This allows a meal recommendation system to present more useful information in an end-to-end system implementation and reason across modalities as well as with content and preparation processes.

The closest to our work is the \textit{SousChef} system~\cite{ribeiro2017souschef}, which tackles the problem of food recommendation\footnote{They call it meals but do not have the rich meal configurations we support, and instead focus on single items.} for improving the health of older adults. \textit{SousChef} utilizes a two-stage rule-based algorithm to filter away incompatible choices and then uses user preference data and item ingredients to recommend multi-item food plans based on nutritional needs; we consider a longer horizon problem setting, learning-based methods, and a multi-modal dataset. \textit{Eat This Much}~\cite{eatthismuch} is another implementation that considers the problem of meal recommendation and provides users with the options to input calories, diet preference(s), and the number of meals they would like. However, this work does not disclose the recommendation algorithm, nor does it allow for the customization of meal configurations and time horizons that our work does. Refer to Table~\ref{tab:qual_comparison} for a comparison of our work with the aforementioned two works.
Additionally, other systems~\cite{zeevi2015personalized, forouzandeh2024health, min2019food, zioutos2023healthy, pecune2020recommender} have explored various aspects of food recommendation, but often do not incorporate a structured representation that includes both the food content and preparation process.


\begin{table}[]
\begin{tabular}{@{}llll@{}}
\toprule
System Feature        & \textbf{BEACON} & SousChef & EatThisMuch \\ \midrule
Var. Time Horizon & \multicolumn{1}{c}{$\checkmark$}   &          &               \\
Var. Meal Count   & \multicolumn{1}{c}{$\checkmark$}   &          & \multicolumn{1}{c}{$\checkmark$}       \\
Var. Meal Config. & \multicolumn{1}{c}{$\checkmark$}   &          &               \\
Goodness Score(s)        & \multicolumn{1}{c}{$\checkmark$}  & \multicolumn{1}{c}{$\checkmark$}   &         \\     
\hline
\end{tabular}
\caption{Qualitative comparison of BEACON  with related food/ meal systems. (Var. $\rightarrow$ Variable)}
\label{tab:qual_comparison}
\end{table}

\section{Problem Formulation} \label{sec:problem}
We define the problem statement by explaining the inputs and outputs of the proposed system (Figure \ref{fig:archi}). Let $\mathcal{R}$ be the set of all recipes in the R3 representation. Let $\mathcal{U}$ be the set that contains all user-provided information, including dietary conditions (Healthy or Diabetic), their ternary food preferences (\textit{likesDairy, likesNuts, likesMeat}), and meal plan format constraints $\mathcal{C}$. $\mathcal{C}$ includes daily meal names corresponding to the number of meals requested per day (such as Mid-Morning Snack or Dinner), the meal components for each represented as a set $MC \subset \{\text{Beverage}, \text{Main}, \text{Side}, \text{Dessert}\}$, and the length of the meal plan in days.
\footnote{In our simulated case study, we configure all recommendations to be in the following format (per day): Breakfast (Main Course, Beverage), Lunch (Main Course, Side, Beverage), and Dinner (Main Course, Side, Dessert, Beverage). The duration for recommendations is configured separately for different simulations (min one day and max five days).}
The output of our system $\mathcal{MP}$ is a meal plan consisting of meals from $\mathcal{R}$ informed by $\mathcal{U}$ in JSON format with a schema defined by $\mathcal{C}$, with meal items outputted by one of three methods, detailed in \textbf{Solution Components} section: random selection, sequential selection, and bandit-based selection.

\section{Solution Components and \\ BEACON System Implementation} \label{sec:system_implementation}

Figure \ref{fig:archi} shows the proposed BEACON's architecture. 
\begin{table}[ht]
{\scriptsize
    \centering
\begin{tabular}{@{}lllll@{}}
\toprule
Category                              & \% with nuts & \% with meat          & \% with dairy & Total Recipes \\ \midrule
McDonald's                            & 9.10         & 63.6                  & 90.0          & 11            \\
TREAT Recipes & 17.2         & 34.5                  & 48.3          & 29            \\
Taco Bell                             & 0            & 60                    & 100           & 10            \\
Soul Food                             & 0            & 0 & 0             & 2             \\ 
\hline
\end{tabular}
    \caption{Percentage of recipes with nuts (\texttt{hasNuts}), with meat (\texttt{hasMeat}), with dairy products (\texttt{hasDairy}), and the total number of recipes under each category.}
    \label{tab:recipe-stats}
}
\end{table}

We will describe the two main components of our work: (1) our data and related experiments, and (2) our recommendation methods and corresponding evaluation metrics

\subsection{Data Component: Recipes and R3 Preparation}

We will first motivate our chosen data representation for our recipes, then our methods for converting new recipes to this representation, and finally, the effectiveness of our approach in converting food in that format. 

 Existing recipes on the internet are available as inconsistently structured textual documents which makes it difficult for machines to read and reason. Better representation of such information can improve decision support systems and also provide an easy way to query and get insights from the data. \cite{pallagani2022rich} introduced a Rich Recipe Representation (R3) which represents recipes in a structured JSON format. They created twenty-five egg-based recipes in R3 manually from original recipes taken from the RecipeQA dataset \cite{yagcioglu2018recipeqa}. 
To expand the existing twenty-five R3 recipes from the RecipeQA dataset, we considered fast food recipes which are known for their convenient access and soul food recipes which are culturally relevant to the African-American population.
In this work, we generate recipes for eleven items served by McDonald's, ten served by Taco Bell, and two commonly known soul food dishes: fried okra and pumpkin soup. 

Since LLMs have been reported to be effective in many natural language and data processing tasks \cite{zhang2024largelanguagemodelsdata}, including automated machine translation \cite{chitale-etal-2024-empirical},  we wanted to study the effectiveness of using LLMs for the task of automating the conversion of online recipe texts to corresponding R3 structures. For this text-to-JSON translation task, we employed two in-context learning-based methods \cite {dong2022survey}. 
We note that there are costlier alternatives line finetuning \cite{escarda2024llms} and designing LLMs-from-scratch \cite{pl-task-training-fm}, which we leave as possible future extensions.



The twenty-five RecipeQA recipes were extracted manually, requiring end-to-end human effort, with a method denoted by $RC_0$. We then employed a semi-automated approach denoted by $RC_1$, using ChatGPT (GPT 3.5) to convert the recipes into their intended R3 structure. Finally, we considered a fully automated approach, denoted by $RC_2$ utilizing \textit{Mixtral-8x7B-Instruct-v0.1}'s \cite{jiang2024mixtralexperts} API available through HuggingFace.

\subsubsection{R3 Evaluation Metrics}
Before expanding on the usage of LLMs for R3 conversion, it is pertinent to discuss the metrics for evaluation of the conversions. We describe four metrics: semantic similarity score ($s_{sem}$), syntatic similarity score ($s_{syn}$), perplexity ($s_{ppl}$), and JSON decoding error count ($s_{jec}$).

\noindent\textbf{Semantic Similarity Score}: To assess if the semantic meaning and essence is preserved in the R3 representation, we generate an embedding vector for the original recipe text ($v_{org}$) and for the R3 JSON string ($v_{r3}$), and calculate the BERTScore \cite{zhang2020bertscoreevaluatingtextgeneration} $s_{sem} \in [-1, 1]$.

\noindent\textbf{Syntatic Similarity Score}: To assess if the structure of the R3 representation of a metric is similar to other R3 representations in terms of the present JSON keys, we initialize a set, $R3_{keys} = \{(k, n)\}$, where $k$ is a key in the JSON string and $n$ is the nesting level of the key. We then populate $R3_{keys}$ with all of the keys and nesting levels in the R3 representations of the twenty-five egg-based recipes curated using method $RC_0$.

\noindent\textbf{Perplexity}: As it is common to measure the informativeness of an LLM's output, we use perplexity to assess the confidence of the LLM's generated R3 representations \cite{miaschi-etal-2021-makes-perplexity}. 
We denote it by $s_{ppl}$.

\noindent \textbf{JSON Decode Error Count}: LLMs tend to make mistakes when generating a JSON string, including missing and extra characters like quotations, brackets, and commas. For this reason, when processing an LLM's output, we count the number of these mistakes, $s_{jec}$.

\subsubsection{LLM-based  Conversion Methods} 

We now present two methods for recipe format conversion, one semi-automated ($RC_1$) and another automated ($RC_2$).

\noindent\textbf{Hybrid-LLM conversion of recipes to R3} ($RC_1$): Converting recipes with this approach exploits the summarization and machine translation in LLMs and involves manually collating different portions of the R3 representation, as described below.

\noindent\textbf{(1) Ingredients and Nutrients Extraction:} For extracting ingredients with amounts and units, we use 0-shot prompting. We copied the list of ingredients with their amounts as presented in the online recipes and input it into ChatGPT with the prompt ``Please extract the ingredients from this list in this JSON structure: ", with the JSON structure in Figure \ref{fig:ingredient_json} appended. We provided a similar prompt and JSON structure (Figure \ref{fig:macronutrient_json} for nutritional information, including macronutrients, such as proteins or carbohydrates,  and micronutrients like vitamins and minerals.

\noindent \textbf{(2) Instruction Extraction}: For properly extracting instructions with atomic steps, we utilize the chain-of-thought prompting and few-shot prompting~\cite{ma2024fairness, zhang2023machine} by first inputting to it the prompt seen in Figure \ref{fig:instruction_extraction}. Following this, we provided it a series of prompts with JSON strings of the form seen in Figure \ref{fig:instruction_json}.

\noindent \textbf{(3)~Meal Component Annotation}: Each of the input recipes are structured in a formal representation using JSON and contain individual food items that need to be categorized based on their roles in the meal. This includes ``Beverage," ``Main Course," ``Side," and ``Dessert." Given that these roles can vary depending on context and cultural interpretations, the task can easily be subjective (and inconsistent) to do it manually. For example, an item such as "bread" may be a side dish in one context but as part of the main course in another. To handle such ambiguities, the annotation task makes use of an LLM to analyze the recipe and assign appropriate roles to each food item. More than one food role for each item is possible. We chose ChatGPT-3.5 as our annotator. The prompt, seen in Figure \ref{fig:food_item_annotator}, asks the LLM to consider possible contextual relationships within an input recipe and interpret the necessary nuances for it to map food roles. 


 In addition to the aforementioned LLM-assisted portions, we manually annotate each item's R3 representation with its binary food features (\texttt{hasDairy}, \texttt{hasNuts}, \texttt{hasMeat}). Table \ref{tab:recipe-stats} shows the \% of recipes with nuts, meat, and dairy products, and the total number of recipes under each category.
We use $RC_1$ for
twenty-one fast food, two soul food, and four general recipes (twenty-seven total) 
combine them with twenty-five from \cite{pallagani2022rich} to have a total of fifty-two recipes in R3 format. 



\noindent \textbf{Fully-automated LLM conversion of recipes to R3 ($RC_2$)}

Noting the manual effort required in $RC_0$ and $RC_1$,
in $RC_2$, we were motivated to study if LLMs could be used to generate the entire R3 JSON structure of a given recipe. 
We opted to use \textit{Mixtral-8x7b-Instruct-v0.1}, as it is freely available for inference through the Hugging Face inference API and offers a token context length large enough to encapsulate entire recipe texts and R3 examples in its free tier. We studied three main variables in our experiments: temperature, number of examples provided, and atomicity of examples, number of examples provided. Temperature is well-studied LLM parameter in recent literature \cite{peeperkorn2024temperaturecreativityparameterlarge}, and we consider values of $t \in \{0.0, 0.3, 0.7, 1.0\}$. Since few-shot prompting has been an effective method to extract relevant output from LLMs, we provided examples of structures within the R3 representation to the LLM within the system prompt.
We consider two types of examples, the entire recipe in R3 ($e_1$) and the cooking instructions part of recipe in R3 (Figure~\ref{fig:instruction_json}), $e_2$.
We consider three variants of few-shot prompting with 1-shot, 3-shot, and 5-shot. With four values for temperature, three values for the number of examples, and two values for the types of examples provided, we determine which of twenty-four possible configurations of variables would be most effective in generating R3 representations. The example type and the number of examples are encapsulated within the system prompt (Figure \ref{fig:llm_sys_prompt}). We then provide the prompt shown in Figure~\ref{fig:llm_user_prompt} with the recipe text inserted, and record the model's attempt at producing the corresponding R3 representation.

We evaluate the performance of \textit{Mixtral-8x7b-Instruct-v0.1} in generating R3 representations of 5 recipes across these 24 configurations and present them in Figure \ref{tab:llm_experiments}. For brevity, we refer to providing full recipes as examples as $e_1$ and instructions as examples as $e_2$. {\bf We found that a single configuration (1-shot, $e_1$, temp of 0.3) consistently returns a valid JSON conversion}, so we selected it as our representative converter for the $RC_2$ method.
To ascertain which of the three methods ($RC_0$, $RC_1$, $RC_2$) is most effective in converting recipe text to R3, we consider three sets: the twenty-five R3 representations curated with $RC_0$ to gauge $RC_0$, the twenty-seven R3 representations curated with $RC_1$ to gauge $RC_1$, and the fifty-two original recipe texts of the curated R3 representations to gauge $RC_2$. The results of applying the aforementioned metrics can be found in Figure \ref{tab:r3_conv_exp}.  


\begin{table*}[t]
\centering
\makebox[\textwidth]{ 
\begin{tabular}{@{}lllll|lllll|lllll@{}}
    \toprule
    \multicolumn{5}{c|}{1-shot prompting} & \multicolumn{5}{c|}{3-shot prompting} & \multicolumn{5}{c}{5-shot prompting} \\
    \midrule
    \textbf{Config} & \textbf{$s_{jec}$} & \textbf{$s_{syn}$} & \textbf{$s_{sem}$} & \textbf{$s_{ppl}$} & 
    \textbf{Config} & \textbf{$s_{jec}$} & \textbf{$s_{syn}$} & \textbf{$s_{sem}$} & \textbf{$s_{ppl}$} & 
    \textbf{Config} & \textbf{$s_{jec}$} & \textbf{$s_{syn}$} & \textbf{$s_{sem}$} & \textbf{$s_{ppl}$} \\
    $e_1$, $t=0.0$ & 437 & - & 0.666 & \textbf{1.06} & 
    $e_1$, $t=0.0$ & 86 & - & 0.640 & 1.06 & 
    $e_1$, $t=0.0$ & 93 & - & \textbf{0.766} & 1.03 \\
    $e_1$, $t=0.3$ & 59 & \textbf{0.361} & 0.630 & 1.24 & 
    $e_1$, $t=0.3$ & 212 & - & 0.677 & \textbf{1.03} & 
    $e_1$, $t=0.3$ & 690 & - & 0.516 & \textbf{1.02} \\
    $e_1$, $t=0.7$ & 84 & - & 0.681 & 2.32 & 
    $e_1$, $t=0.7$ & 52 & - & 0.629 & 15.6 & 
    $e_1$, $t=0.7$ & 86 & - & 0.667 & 2.98 \\
    $e_1$, $t=1.0$ & 28 & - & \textbf{0.706} & 47.6 & 
    $e_1$, $t=1.0$ & 111 & - & 0.686 & 49.4 & 
    $e_1$, $t=1.0$ & 54 & - & 0.662 & 52.6 \\
    
    $e_2$, $t=0.0$ & 288 & - & 0.639 & 1.09 & 
    $e_2$, $t=0.0$ & 221 & - & 0.700 & 1.06 & 
    $e_2$, $t=0.0$ & 1283 & - & 0.701 & 1.04 \\
    $e_2$, $t=0.3$ & 387 & - & 0.639 & 1.13 & 
    $e_2$, $t=0.3$ & 65 & - & 0.641 & 1.26 & 
    $e_2$, $t=0.3$ & \textbf{28} & - & 0.717 & \textbf{1.02} \\
    $e_2$, $t=0.7$ & \textbf{41} & - & 0.688 & 3.29 & 
    $e_2$, $t=0.7$ & 187 & - & 0.667 & 2.27 & 
    $e_2$, $t=0.7$ & 578 & - & 0.604 & 1.56 \\
    $e_2$, $t=1.0$ & 133 & - & 0.689 & 20.1 &
    $e_2$, $t=1.0$ & \textbf{40} & - & \textbf{0.766} & 46.5 & 
    $e_2$, $t=1.0$ & 95 & - & 0.735 & 48.8 \\
    \bottomrule
\end{tabular}
}
\caption{R3 Conversion Experiments using $RC_2$ Method. Bolded values indicate the best-performing method on that metric.}
\label{tab:llm_experiments}
\end{table*}

\begin{table}[]
\begin{tabular}{@{}ccccc@{}}
\toprule
\textbf{Config} & $s_{jec}$ & $s_{syn}$ & $s_{sem}$ & $s_{ppl}$ \\ \midrule
$RC_0$             & 0    & 0.584          & \textbf{0.763} & -    \\
$RC_1$             & 0    & \textbf{0.778} & 0.733          & -    \\
$RC_2\alpha$        & 26   & 0.376          & 0.758          & 1.02 \\
$RC_2\beta$         & 1    & 0.238          & 0.715          & 1.27 \\ \bottomrule
\end{tabular}
\caption{R3 Conversion Methods Comparison. Bolded values indicate the best-performing method on that metric. Note that $RC_2\alpha$ is the best-performing configuration from \ref{tab:llm_experiments} and $RC_2\beta$ is the second best.}
\label{tab:r3_conv_exp}
\end{table}


\subsection{Meal Recommendations Component}

We will first present the goodness metrics for evaluation meal plans, then recommendation methods, and finally the evaluation for said recommendation methods.

\subsubsection{Goodness Metrics for Recommendations}

We use three different evaluation criteria to evaluate a recommendation. These criteria include assessing the recurrence of items across meals and within meals (duplicate metric, $md$), assessing if the recommended meals satisfy the preferred components (meal coverage metric, $cs$), and assessing if the recommended meals have ingredients that match user preferences (user-constraint metric, $uc$). 
For each recommendation, we compute the goodness score $G$ as a weighted sum of the individual scores $md$, 
$cs$, and $uc$, with weights tailored to user preferences (e.g., some users may prefer having duplicates in their meals, while others may not).
We elaborate on the three metrics below.


\noindent \textbf{Duplicate Metric ($dm$)}
Our duplicate metric examines the occurrence of repeated items in meals as a \textit{meal item duplicate}. The meal item duplicate score, denoted as $dm_i$ for a particular meal $m_i$, measures the ratio of unique items to total items within the meal. For a recommendation comprising multiple meals $m_1, ..., m_n$, we calculate the meal item duplicate score $dm$ as the average of all meal item duplicate scores $dm_1, ..., dm_n$. 

\noindent \textbf{Meal Coverage Metric ($mc$)} Our meal coverage metric evaluates the extent to which a meal recommendation aligns with the user's desired food roles (Main Course, Side Dish, Dessert, or Beverage). To calculate the coverage score for a recommendation, we analyze the presence of recommended meal items corresponding to their role and alignment with user preferences. If a recommended item matches its assigned role and aligns with user preferences (indicated using a weight of $+1$), it positively contributes to the coverage score. Conversely, misaligned recommended items (e.g., recommending a beverage item like soda as a side dish) would incur a penalty on the coverage score. For each meal, $m_i$, we calculate a coverage score $mc_i$ by taking the ratio of requested roles fulfilled to the number of requested roles. We calculate our final coverage score, $mc$, as the average of all scores $mc_1, ..., mc_n$.

\noindent\textbf{User Constraint Metric ($uc$)} In addition to specifying the types of food roles they prefer in each meal, users also provide their ingredient preferences, focusing on three key features: dairy content, meat content, and nut content. These features were chosen to create a minimally functional system, with plans to extend the list of features in the future. Each feature can have a user preference value of $-1$, $0$, or $+1$, representing a negative preference (the user prefers meals without this ingredient), a neutral preference (the user is indifferent), and a positive preference (the user prefers meals with this ingredient), respectively. Our system is designed to be flexible, allowing any number of such features to be added or removed based on user requirements. 

We manually annotate each of our fifty-two R3 representations with corresponding feature flags, indicating whether an item contains a particular ingredient (e.g., dairy, meat, or nuts). For each meal $m_i$, we calculate a user constraint score $uc_i$ by comparing the user's preference with the meal's ingredient content. If the user's preference is negative ($-1$) and the meal contains the ingredient, this counts negatively towards the score. Conversely, if the user's preference is positive ($+1$) and the meal contains the ingredient, this counts positively towards the score. Neutral preferences ($0$) do not affect the score regardless of the meal's content. Additionally, if a meal does not contain an ingredient that the user positively prefers, the overall goodness is not penalized unless the user specifies it through a configurable flag. We calculate the final user constraint score \( uc \) as the average of $uc_1, uc_2, ..., uc_n$ across all meals in the recommendation.



\vspace{-0.7em}

\subsubsection{Recommendation Methods}

We used three different methods to recommend meals to users: (1) \textit{M0: Random/Baseline}, (2) \textit{M1: Sequential}, and (3) \textit{M2: Relational Boosted Bandit}.

\textit{M0} serves as our baseline method for meal recommendations. In this approach, meals are generated randomly, i.e., the selection of items for each meal does not consider any user preferences, dietary restrictions, or allergen information. An advantage of \textit{M0} is that it has the potential to introduce users to a diverse array of food choices. By forming meals without any specific criteria, users might encounter new and varied food items that they may not have otherwise considered, thereby broadening their culinary experiences.

\textit{M1} introduces a more structured approach compared to \textit{M0}. In this method, we use our dataset of recipes and rotate through them to recommend meals. This sequential nature ensures that each recipe in the list is eventually recommended; however, like \textit{M0}, \textit{M1} does not take into account any individual user needs. Its primary advantage over \textit{M0} is only the avoidance of repetitive randomness.

We explain the details of the boosted bandit algorithm used to generate recommendations in the appendix.

\section{Use Case Walk-through and Demonstration} 
\label{sec:demo}


\begin{figure*}[htbp]
\centering
\includegraphics[width=0.49\linewidth]{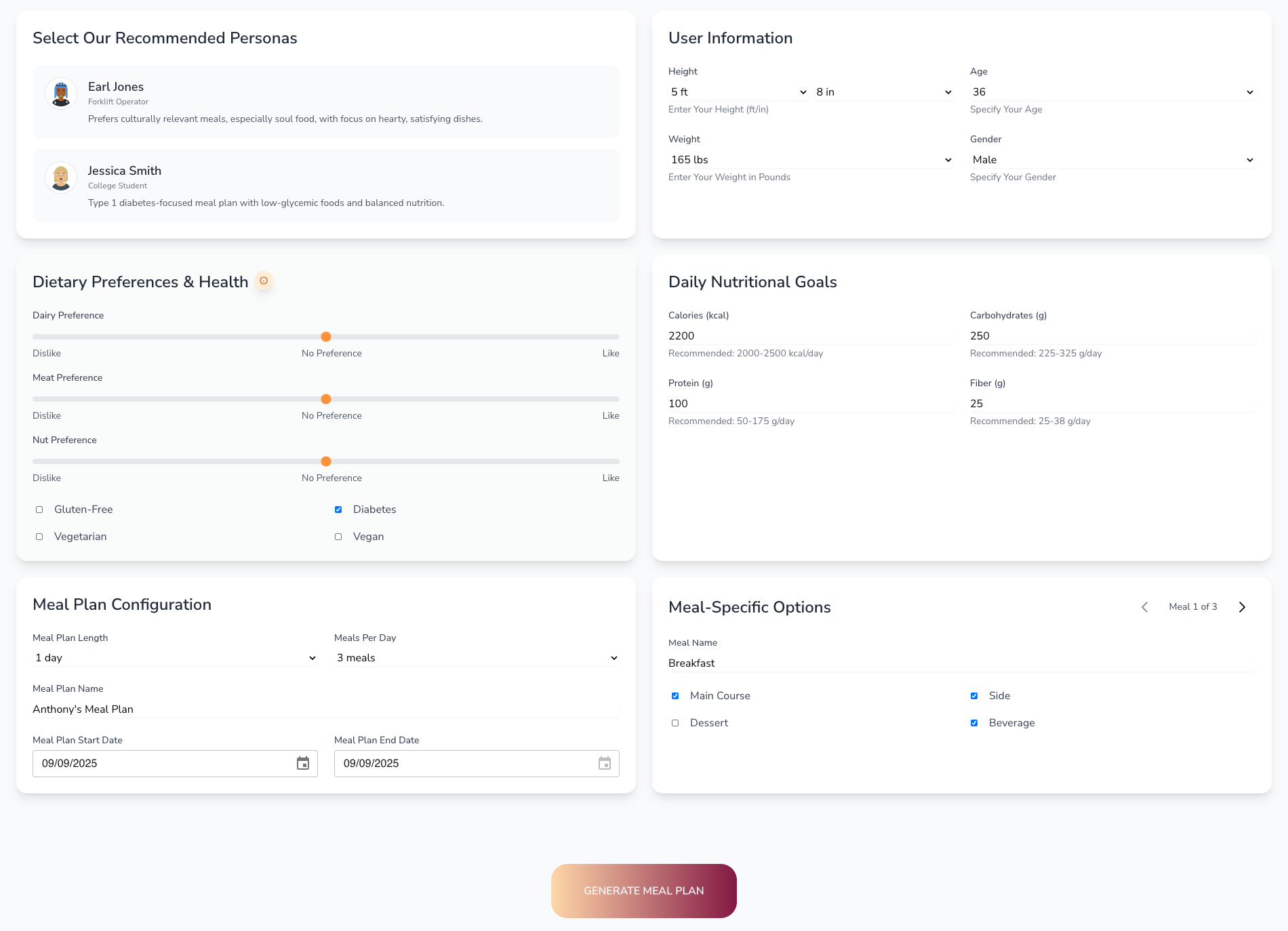} 
\includegraphics[width=0.49\linewidth]{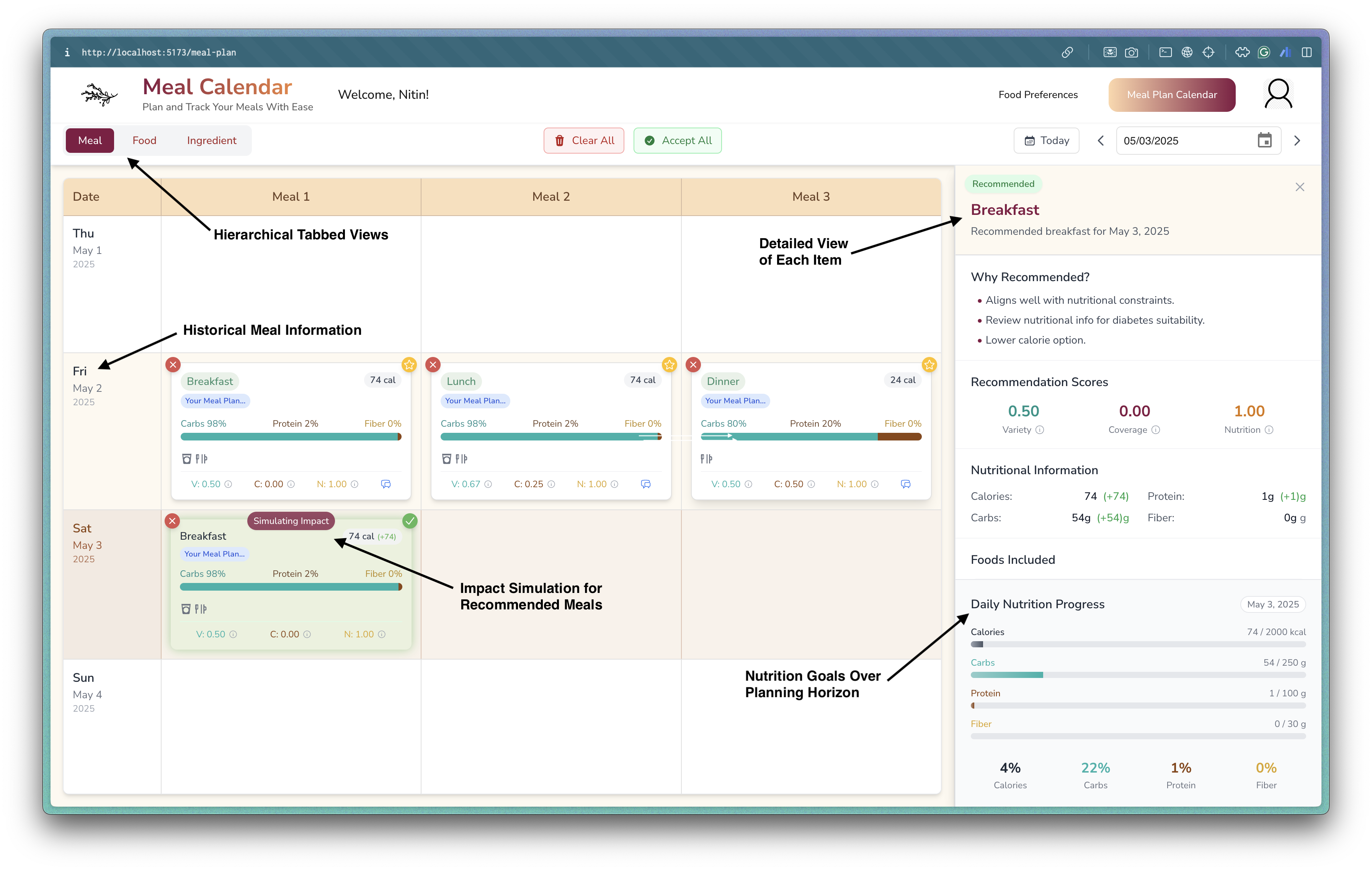}
\caption{
    Left: Meal preference configuration showing Anthony Gibbson's selected constraints, including demographic information, dietary restrictions, nutritional goals, and meal type selection.
    Right: The interactive, calendar-based meal visualization for Anthony Gibbson, tailored to his preferences and health conditions. The system allows Anthony to build, visualize, realize, and track his meal plans. Arrows and annotated text are presented for clearer explanations.
}
\label{fig:boh_screens}
\label{fig:boh_screens}
\end{figure*}


In the following sections, we describe a detailed use case of the BEACON system and consider the ethical, legal, and security ramifications of the deployed system. Our proposed solution can be used in a variety of different use cases, including serving as a meal planner for
(a) diabetic individuals looking to be recommended meals that help them manage their condition,
(b) individuals from minority communities looking to be recommended culturally relevant meals,
(c) busy professionals seeking convenient and healthy meal options, and
(d) medical professionals seeking to recommend meal plans to their patients.



\subsection{Use Case: Managing Diabetes}




This walk-through highlights the two main interfaces of the system: \textbf{Meal Preference Configuration} and \textbf{Meal Calendar Presentation}, as shown in Figure~\ref{fig:boh_screens}.
To illustrate the system’s functionality, we consider a fictional \textit{persona}: Anthony Gibbson, an African-American individual managing diabetes. Anthony seeks meal plans that align with his dietary needs and cultural food preferences. Through our application, Anthony is guided to configure his profile, receive tailored meal plans, and monitor his dietary patterns with actionable insights. Detailed figures of the system are given in the appendix.

\subsubsection{Meal Preference Configuration}

Anthony begins his journey on the Food Preferences page, the interactive interface depicted in Figure~\ref{fig:boh_screens} left. Here, he has the option to select a predefined persona that populates the page with default values or, as in this walk-through, to enter his information manually for a fully personalized experience.

He populates the form with his anthropometric data (such as height and weight), dietary preferences, and specific nutritional goals. For his dietary preferences, he specifies `diabetes' as his primary health condition and, since he has no specific restrictions regarding dairy, meat, or nuts, Anthony leaves the corresponding sliders unchanged. Additionally, he opts not to select a predefined persona, such as a college student or a worklift operator. To further tailor the plan, he sets his nutritional goals, like limiting daily carbohydrate intake, and configures meal-specific options, such as meal names and times.

The BEACON algorithm processes these inputs, incorporating demographic and health constraints to generate a personalized meal plan. While this example focuses on Anthony, the system is designed for a diverse range of users with varying health conditions and cultural preferences.



\subsubsection{Personalized Meal Calendar Presentation}  




%
Once Anthony submits his preferences, the system presents him with a tailored meal plan via the interactive Meal Calendar, as shown in Figure~\ref{fig:boh_screens} right. This interface moves beyond static lists, offering a dynamic environment where Anthony can build, visualize, realize, and track his meal plans over time. This interface is designed with multiple views to give him full control over how he visualizes and interacts with his plan.

By default, he is presented with the \textbf{Meal View}, which shows his recommended meals sequentially. From this view, he can accept a recommendation to save it to his meal history, reject it, favorite it to influence future suggestions, or regenerate all recommendations for a fresh set of options. Clicking on any meal reveals its key nutritional information, such as calories, carbohydrates, protein, and fiber. For more details, Anthony can switch to the \textbf{Food Item View}. This breaks down each meal into its individual components: main courses, side dishes, and beverages. This is particularly useful if he decides to adhere to only part of a recommendation. For instance, he can click on `waffles' to view its specific nutritional data and access the recipe needed to prepare it. Finally, the \textbf{Ingredient View} provides the most granular level of detail by breaking down each food item into its constituent ingredients. This view serves as a practical tool, giving Anthony a quick glance at all the ingredients he will need to purchase to prepare his scheduled meals for the week.

The system also provides actionable insights and nudges to help him adhere to his goals for diabetic management. For instance, if it detects that he schedules high-sugar meals for several consecutive days, it might visually flag those entries and suggest healthier, culturally relevant alternatives. This behavioral feedback loop learns from his choices to refine future recommendations, ensuring they are both effective and sustainable.

These views empower Anthony to interact with his plan in several ways: (i) Build and Replan: He can dynamically adjust the plan by manually editing meals, accepting AI-driven suggestions, or regenerating specific meals of entire days if his needs change. (ii) Visualize and Track Goals: The calendar displays his progress towards nutritional targets (e.g., daily carbohydrate limits). Visual cues, such as progress bars or alerts, highlight how each meal impacts his goals. (iii) Realize the Meal Plan: For any given meal, he can choose how to execute it. He can access cooking instructions or opt to order the necessary meal through integrated services. (iv) Analyze Historical Patterns: The system allows him to review past consumption data, helping him recognize recurring habits, such as frequently skipping lunch or relying on the same side dishes, and make informed adjustments.

This intuitive dashboard transforms the meal plan into a comprehensive dietary partner, empowering Anthony to evaluate, adjust, and realize his health objectives.






\subsection{Ethical, Legal, and Security Considerations}

Ensuring ethical, legal, and secure practices is central to the BEACON system. The system is distributed under the MIT license, ensuring easy access while crediting appropriate intellectual property (IP).

User privacy is prioritized by enabling access to core functionality without requiring personal data. Optional information, such as dietary preferences, is securely stored in a protected database solely for personalizing recommendations. Users have full control over their data, with the option to delete their account at any time, ensuring compliance with privacy standards like GDPR \cite{gdpr}. The system also aligns with the NIST Risk Management Framework (RMF) \cite{nist_rmf} and California AI risk guidelines, including the California Consumer Privacy Act (CCPA) \cite{ccpa}, which emphasize harm mitigation and data privacy.
Bias and discrimination risks are mitigated through an inclusive design that avoids reliance on demographic data. Targeted meal plans, such as those for African American users with diabetes, ensure cultural and health relevance. 


To minimize system abuse, recipe sharing is disabled by default, and all recipe data is vetted for accuracy and sourced from verified databases such as MyPlate \cite{myplate}. Health disclaimers encourage users to consult professionals for dietary advice, further reducing the risks of misinformation. BEACON complies with all applicable technology licenses, including the MIT and BSD-3-Clause licenses. Health data management adheres to HIPAA standards, and demographic data complies with GDPR \cite{gdpr} and CCPA \cite{ccpa}. Intellectual property is managed under the MIT license and university policies, supported by a Contributor License Agreement (CLA) for transparency.

Security measures include encrypting sensitive information, such as login credentials and optional health data, with robust methods like Two-Factor Authentication (2FA). Potential attack vectors, such as Cross-Site Scripting (XSS) and API key misuse, are addressed through input validation, Content Security Policies (CSP), and regular audits.
Finally, scalability is achieved by focusing on usability metrics like system uptime and feature usage rather than collecting user-specific data. Regular load testing ensures reliable performance as user demand increases. By adhering to these principles, BEACON aims to be a trustworthy, inclusive, and secure meal recommendation system.

Refer to the appendix for a detailed description of how we plan to implement the system as a web application.

\section{Evaluation} \label{sec:evaluation}


\subsection{Recipe Translation with LLMs Evaluation}

We conclude from the Tables 
 \ref{tab:llm_experiments} and \ref{tab:r3_conv_exp} that freely accessible, pre-trained LLMs alone can not consistently generate large JSON structures. We note that only one out of 24 configurations (1-shot, $e_1$, $t=0.3$ were able to generate a properly formatted JSON string, which is why all other configurations do not report a syntactic score. Additionally, all LLM-based methods of recipe conversion ($RC_1, RC_2\alpha, RC_2\beta$) consistently score more than the manual $RC_0$ method in terms of preserved semantic information. We also note that manually curated recipes report a significantly lower syntactic score than ones generated using method $RC_1$, which is because we consider the set of all keys in all of our R3 recipes as our reference set. LLMs may produce a lot of unnecessary keys that are not in the manually curated set of R3 recipes. Expectedly, manually curated R3 recipes report the highest semantic score as a human ensures that information is not lost in translation. We conclude that LLMs must be used in conjunction with rule-based or manual methods for data processing or be fine-tuned on the specific problem domain to be more effective.
\begin{figure*}[htbp]
\centering
    \includegraphics[width=1\linewidth]{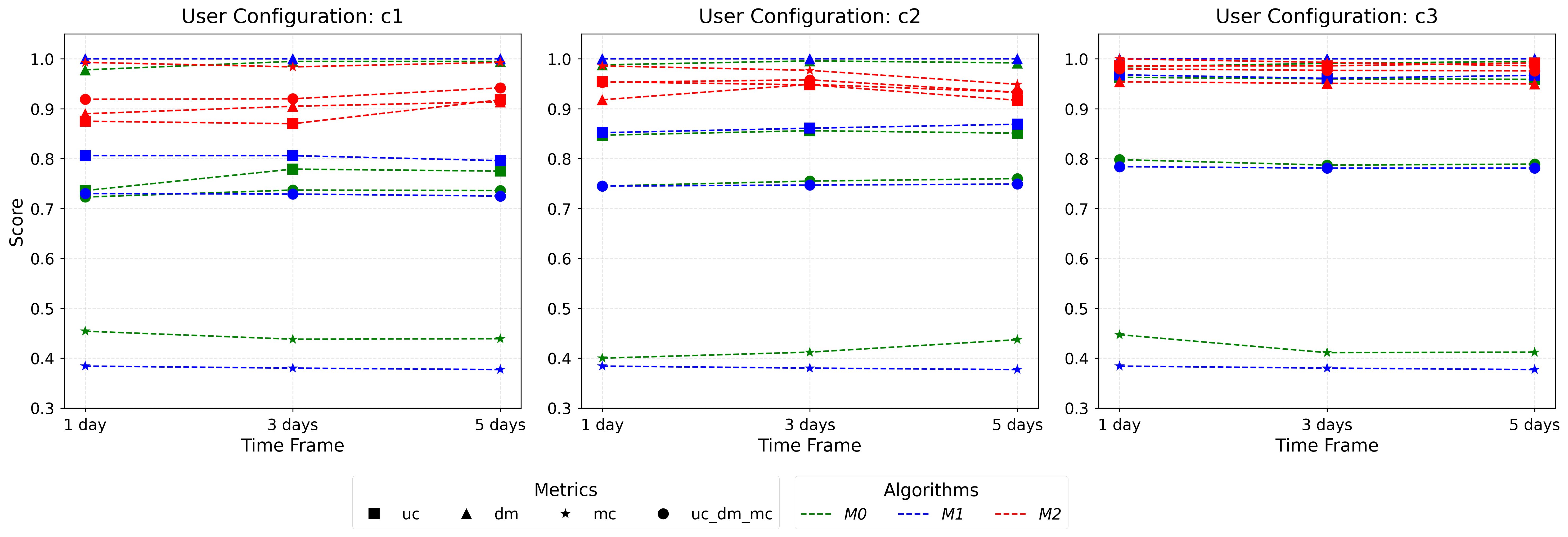}
    \caption{Performance of meal recommendation methods across configurations, time frames, and metrics. $c_1$: 12/0/12, $c_2$: 8/8/8, $c_3$: 2/20/2 (representing decreasing user constraints).}
    \label{fig:reco-charts}
\end{figure*}

\subsection{BEACON Recommendation Evaluation}

\noindent We conduct our experiments for each recommendation method across 3 different time frames $t_1$~(1 day) $t_2$~(3 days), and $t_3$~(5 days) and 3 user configurations, $c_1, c_2,$ and $c_3$. These user configurations all include 24 users. For each of the food features that we consider (\texttt{hasDairy}, \texttt{hasMeat}, \texttt{hasNuts}), we consider the corresponding user features $\{u_f\}$
that can each take on a value of -1, 0, or 1. These values correspond to negative preference, neutral preference, and positive preference respectively. For each $c_i$ and feature $u_f$, we randomly select $p_i$ users to have a positive preference to $u_f$, $n_i$ users to have a negative preference to $u_f$, and the remaining to have a neutral preference to $u_f$. In $c_1$, $c_2$, and $c_3$, we choose $n_1 = p_1 = 12$, $n_2 = p_2 = 8$, and $n_3 = p_3 = 2$. Thus, the constraints on users' preferences are decreasing across the configurations.
Each $c_i$ is referred to as $n_i$/(24 - $n_i$ - $p_i$)/$p_i$ in Figure~\ref{fig:reco-charts}, corresponding to negative, neutral, and positive preference. For each experiment that we conducted, we display our three metrics: user constraint ($uc$), duplicate meal ($dm$), and meal coverage ($mc$), as well as their average ($uc\_dm\_mc$). We display our results graphically in Figure~\ref{fig:reco-charts}. Please refer to Table~\ref{tab:performance_metrics} for an equivalent tabular view.

As shown in Figure~\ref{fig:reco-charts}, \textit{M2} expectedly outperforms other methods in the user constraint metric and meal coverage metric as it is the most informed out of the three. It is important to note that when there are fewer users with negative preferences towards the food features, \textit{M0} and \textit{M1} only perform marginally worse because users are less particular about their preferences. \textit{M1} always scores perfectly in the duplicate meal metric because there are more items in our dataset than are in a meal, while \textit{M2} performs the worst, which is caused by \textit{M2} favoring very few items with a higher probability of being a positively recommended item. This causes \textit{M2} to perform poorly in the $uc\_dm$ combination metric for most trials. Additionally, in the $uc\_dm$ metric, \textit{M2} performs the worst in the $c_3$ configuration because users are less particular and \textit{M2} is more likely to recommend duplicates. However, since, \textit{M0} and \textit{M1} do not perform nearly as well in the $mc$ and $uc$ metrics, \textit{M2} performs significantly better in the $uc\_dm\_mc$, $uc\_mc$, and $dm\_mc$ combination metrics. We can also see that \textit{M2} is a precise and accurate method regardless of the user configuration and \textit{M0} and \textit{M1} are somewhat accurate for some metrics, but lack precision. We thus conclude that the boosted bandit algorithm represented by \textit{M2} is the most effective in recommending meals out of the three.\\

\section{Discussion and Conclusion} \label{sec:discussion}
As mentioned previously, we only considered two LLMs (ChatGPT, \textit{Mixtral-8x7B-Instruct-v0.1}) in our recipe conversion efforts, as we wanted to determine if such an effort could be undertaken at minimal cost. We conclude that these LLMs alone can not successfully translate plain text recipes into R3 representations. A future extension of this work, it may be worthwhile to consider fine-tuning an LLM on recipe translation or considering larger models behind a paywall.

In conclusion, in this paper, we introduced the novel problem of meal recommendation considering different meal configurations and time horizons,  presented our solution which utilizes the boosted bandit method to address the problem of meal recommendation, displayed a dataset of 50 R3 items consisting of non-fast food and fast food items (Taco Bell and McDonald's), contributed a unique goodness metric that can be used to assess the quality of recommendations, showed the efficacy of the boosted bandit method for generating robust recommendations across three user configurations and three time frames, as well as motivating a use case of the BEACON system, which is in development. We believe this can be a promising path toward promoting user adherence to dietary nutrition guidelines while balancing convenience.

In the future, one can extend this work in many directions, including (1) increasing the size of our dataset as this leads to more robust models by exploring more LLM-based approaches; (2) increasing the number of features in terms of ingredients/allergens so that our dataset is more varied, and users with more allergens can receive positive recommendations; (3) experimenting with different recommendation algorithms and methods so that we may further explore the use of R3 representations; and (4) conducting qualitative evaluation to show the acceptance of our recommendation system. 

\section{Acknowledgements}

We thank Prof. Jose Vidal for mentoring the undergraduate students (VN, ZA, AD, NG) on software engineering principles for the Capstone project under whose aegis, the BEACON software is being implemented based on research in Prof. B. Srivastava's group. We also thank Mr. Lokesh Johri, Tantiv4 and MyMealRx.ai for driving discussions to generalize this work beyond the presented scope and use cases. We acknowledge funding support for this work from SCRA and Univ. of South Carolina.

\bibliography{aaai25}
\clearpage
\appendix

\section{Supplementary Figures} \label{appendix:start}

\begin{figure}[htbp]
\begin{lstlisting}[language=json,firstnumber=1]
{
  "name": "Shredded Cheese",
  "quantity": {
    "measure": "2",
    "unit": "cups"
  },
  "allergies": {
    "id": "0x0B76",
    "category": [
      "dairy"
    ],
    "ref": [],
    "details": ""
  },
  "alternative": "",
  "quality_characteristic": "",
  "image": ""
}
\end{lstlisting}
\caption{Representing ingredients in R3}
\label{fig:ingredient_json}
\end{figure}

\begin{figure}[htbp]
\begin{lstlisting}[language=json,firstnumber=1]
{
  "Calories": {
    "measure": "417",
    "unit": "kcal"
   }
}
\end{lstlisting}
\caption{Representing nutrition in R3}
\label{fig:macronutrient_json}
\end{figure}

\begin{figure}[htbp]
\begin{lstlisting}[language=json,firstnumber=1]
{
  "original_text": "Beat the egg well. Egg drop soup is noted for its strands of shredded egg. To achieve this characteristic look and texture, make sure you blend the egg mixture well, and stir it slowly into the broth.",
  "input_condition": [
    "have_egg",
    "have_chicken_broth"
  ],
  "task": [
    {
      "action_name": "Beat the egg well",
      "output_quality": [
        "Make sure it is free from shredded egg strands.",
        "Egg mixture should be well blended."
      ],
      "background_knowledge": {
        "tool": [
          "For beating egg"
        ],
        "failure": [
          "Eggshell fell into the bowl"
        ]
      }
    }
  ]
}
\end{lstlisting}
\caption{Cooking instruction represented as task in R3}
\label{fig:instruction_json}
\end{figure}

\begin{figure}[htbp]
\begin{tcolorbox}
[colback=brown!5!white, colframe=brown!95!black, title=Instruction Extractor: Prompting Task, sharp corners=south]
I am going to provide you with a series of examples of JSON representations of recipe instructions, and I want you to learn the representation such that when I give you a plain-text recipe instruction, I would like you to provide the corresponding JSON representation. Do not provide any other details.
\end{tcolorbox}
\caption{An LLM user prompt for extracting instructions in a JSON structure}
\label{fig:instruction_extraction}
\end{figure}

\begin{figure}[htbp]
\begin{tcolorbox}
[colback=brown!5!white, colframe=brown!95!black, title=Meal Component Annotator: Prompting Task, sharp corners=south]
You are an intelligent food item annotator. Out of the following food roles, [`Beverage', `Main Course', `Side', `Dessert'], your task is to assign the relevant food roles necessary, up to twenty recipes at a time. 

\textbf{Input:}
\begin{itemize}
    \item A set of recipes, each structured into a formal recipe representation in JSON.
\end{itemize}

\textbf{Output:}
\begin{itemize}
    \item Based on the set of recipes given, assign each item the relevant food role(s). 
\end{itemize}
Do not provide any other details. Print the set of recipes in the same order and JSON code structure you were given. You can assign more than one food role if needed.
\end{tcolorbox}
\caption{An LLM user prompt for annotating recipes with meal components}
\label{fig:food_item_annotator}
\end{figure}

\begin{figure}[]
\begin{tcolorbox}
[colback=brown!5!white, colframe=brown!95!black, title=LLM System Prompt, sharp corners=south]
You are an expert in processing recipe data into structured formats. Your task is to transform freeform recipe text into a JSON structure that captures essential information about the recipe. The final JSON should contain the following fields:
\textbf{A description of the schema has been omitted for space restrictions, but can be found in \cite{pallagani2022rich}}

Your task is to extract the above fields from a recipe text and return them in the following JSON format. Ensure that you capture all relevant details from the recipe text and break down instructions into atomic tasks as needed. I will assume that you can extract everything but the instructions with no issue, so I will provide you with detailed breakdowns of the 

---

Abridged Example Output:
\{
      ``recipe\_name": "",
      ``macronutrients": \{\},
      ``food\_role": [],
      ``ingredients": [],
      ``hasDairy": ,
      ``hasMeat": ,
      ``hasNuts": ,
      ``prep\_time": ,
      ``cook\_time": ,
      ``serves": ,
      ``instructions": [
        \{
          ``original\_text": ``Heat a skillet over medium-high heat. Add olive oil and swirl to coat the pan. Add the ground beef and break apart. Once halfway cooked, stir in the taco seasoning. Remove from heat and place into a bowl.,"
          ``input\_condition": [
            ``oven\_preheated",
            ``have\_olive\_oil",
            ``have\_ground\_beef",
            ``have\_chilli\_powder",
            ``have\_cumin",
            ``have\_smoked\_paprika",
            ``have\_coriander",
            ``have\_garlic\_powder",
            ``have\_minced\_onion",
            ``have\_sugar"
          ],
          ``tasks": [
            \{
              ``action\_name": ``Heat skillet",
              ``output\_quality": [
                ``Pan should be over medium-high heat."
              ],
              ``background\_knowledge": \{
                ``tool": [
                  ``Skillet"
                ],
                ``failure": []
              \}
            \},
            \{
              ``action\_name": ``Add olive oil and swirl",
              ``output\_quality": [
                "Coat the pan evenly."
              ],
              ``background\_knowledge": \{
                ``tool": [
                  ``Olive oil"
                ],
                ``failure": []
              \}
            \},
            \{
              ``action\_name": ``Add ground beef",
              ``output\_quality": [
                ``Break apart the beef."
              ],
              ``background\_knowledge": \{
                ``tool": [],
                ``failure": [
                  ``Overcooking beef",
                  ``Undercooking beef"
                ]
              \}
            \},
            \{
              ``action\_name": ``Stir in taco seasoning",
              ``output\_quality": [
                ``Ensure even distribution of seasoning"
              ],
              ``background\_knowledge": \{
                ``tool": [],
                ``failure": [
                  ``Spices not evenly mixed"
                ]
              \}
            \},
            \{
              ``action\_name": `Remove from heat",
              ``output\_quality": [
                ``Beef should be halfway cooked."
              ],
              ``background\_knowledge": \{
                ``tool": [],
                ``failure": [
                  ``Overcooking beef",
                  ``Undercooking beef"
                ]
              \}
            \},
            \{
              ``action\_name": ``Place into a bowl",
              ``output\_quality": [],
              ``background\_knowledge": \{
                ``tool": [
                 ``Bowl"
                ],
                ``failure": []
              \}
            \}
          ],
          ``output\_condition": [
            ``have\_cooked\_beef"
          ],
          ``modality": \{
            ``image": [],
            ``video":  
          \}
        \}
      ]
\}

--- 

Your goal is to ensure the JSON format follows the described structure and every detail from the recipe is extracted and formatted correctly. Pay special attention to the breakdown of instructions into atomic tasks, and the conditions, tools, and failure states related to each step.
\end{tcolorbox}
\caption{$RC_2$ system prompt for 1-shot prompting and $e_2$ example type}
\label{fig:llm_sys_prompt}
\end{figure}

\begin{figure}[htbp]
\begin{tcolorbox}
[colback=brown!5!white, colframe=brown!95!black, title=LLM User Prompt, sharp corners=south]
Convert this recipe into R3 format: \textbf{insert recipe plain text}. Please provide only JSON in your response and no backticks or any other text.
\end{tcolorbox}
\caption{$RC_2$ LLM user prompt for recipe translation from plain-text to R3}
\label{fig:llm_user_prompt}
\end{figure}

\newpage

\begin{table*}[t]
\centering
\begin{tabular}{c|c|c|c|c|c|c|c|c}
\toprule
\textbf{Configuration} & \textbf{Algorithm} & \textbf{uc} & \textbf{dm} & \textbf{mc} & \textbf{uc\_dm\_mc} & \textbf{uc\_dm} & \textbf{uc\_mc} & \textbf{dm\_mc} \\ \midrule

$c_1, t_1$ & bandit     & \textbf{0.875} & 0.890 & \textbf{0.993} & \textbf{0.919} & 0.883 & \textbf{0.934} & \textbf{0.942} \\
$c_1, t_1$ & sequential & 0.806 & \textbf{1.000} & 0.384 & 0.730 & \textbf{0.903} & 0.595 & 0.692 \\
$c_1, t_1$ & random     & 0.736 & 0.978 & 0.454 & 0.723 & 0.857 & 0.595 & 0.716 \\ \midrule

$c_1, t_2$ & bandit     & \textbf{0.870} & 0.905 & \textbf{0.984} & \textbf{0.920} & 0.888 & \textbf{0.927} & \textbf{0.944} \\
$c_1, t_2$ & sequential & 0.806 & \textbf{1.000} & 0.380 & 0.729 & \textbf{0.903} & 0.593 & 0.690 \\
$c_1, t_2$ & random     & 0.779 & 0.995 & 0.438 & 0.737 & 0.887 & 0.608 & 0.716 \\ \midrule

$c_1, t_3$ & bandit     & \textbf{0.918} & 0.914 & \textbf{0.993} & \textbf{0.942} & \textbf{0.916} & \textbf{0.955} & \textbf{0.954} \\
$c_1, t_3$ & sequential & 0.796 & \textbf{1.000} & 0.377 & 0.725 & 0.898 & 0.587 & 0.689 \\
$c_1, t_3$ & random     & 0.775 & 0.995 & 0.439 & 0.736 & 0.885 & 0.607 & 0.717 \\ \midrule

$c_2, t_1$ & bandit     & \textbf{0.954} & 0.918 & \textbf{0.986} & \textbf{0.953} & \textbf{0.936} & \textbf{0.970} & \textbf{0.952} \\
$c_2, t_1$ & sequential & 0.852 & \textbf{1.000} & 0.384 & 0.745 & 0.926 & 0.618 & 0.692 \\
$c_2, t_1$ & random     & 0.847 & 0.988 & 0.400 & 0.745 & 0.918 & 0.624 & 0.694 \\ \midrule

$c_2, t_2$ & bandit     & \textbf{0.948} & 0.949 & \textbf{0.977} & \textbf{0.958} & \textbf{0.948} & \textbf{0.962} & \textbf{0.963} \\
$c_2, t_2$ & sequential & 0.861 & \textbf{1.000} & 0.380 & 0.747 & 0.931 & 0.621 & 0.690 \\
$c_2, t_2$ & random     & 0.856 & 0.996 & 0.412 & 0.755 & 0.926 & 0.634 & 0.704 \\ \midrule

$c_2, t_3$ & bandit     & \textbf{0.917} & 0.933 & \textbf{0.949} & \textbf{0.933} & 0.925 & \textbf{0.933} & \textbf{0.941} \\
$c_2, t_3$ & sequential & 0.869 & \textbf{1.000} & 0.377 & 0.749 & \textbf{0.934} & 0.623 & 0.689 \\
$c_2, t_3$ & random     & 0.851 & 0.992 & 0.437 & 0.760 & 0.922 & 0.644 & 0.715 \\ \midrule

$c_3, t_1$ & bandit     & \textbf{0.986} & 0.954 & \textbf{1.000} & \textbf{0.980} & 0.970 & \textbf{0.993} & \textbf{0.977} \\
$c_3, t_1$ & sequential & 0.968 & \textbf{1.000} & 0.384 & 0.784 & \textbf{0.984} & 0.676 & 0.692 \\
$c_3, t_1$ & random     & 0.963 & 0.984 & 0.447 & 0.798 & 0.973 & 0.705 & 0.715 \\ \midrule

$c_3, t_2$ & bandit     & \textbf{0.986} & 0.951 & \textbf{0.993} & \textbf{0.977} & 0.968 & \textbf{0.990} & \textbf{0.972} \\
$c_3, t_2$ & sequential & 0.961 & \textbf{1.000} & 0.380 & 0.781 & \textbf{0.981} & 0.671 & 0.690 \\
$c_3, t_2$ & random     & 0.960 & 0.991 & 0.411 & 0.787 & 0.975 & 0.686 & 0.701 \\ \midrule

$c_3, t_3$ & bandit     & \textbf{0.992} & 0.950 & \textbf{0.986} & \textbf{0.976} & 0.971 & \textbf{0.989} & \textbf{0.968} \\
$c_3, t_3$ & sequential & 0.967 & \textbf{1.000} & 0.377 & 0.781 & \textbf{0.983} & 0.672 & 0.689 \\
$c_3, t_3$ & random     & 0.959 & 0.995 & 0.412 & 0.789 & 0.977 & 0.686 & 0.704 \\ \bottomrule
\end{tabular}
\caption{Performance metrics for different algorithms in BEACON across configurations and timeframes. $c_1$: 12/0/12, $c_2$: 8/8/8, $c_3$: 2/20/2 (representing decreasing user constraints). For each configuration, the best-performing values are in bold.}
\label{tab:performance_metrics}
\end{table*}

\newpage
\section{Boosted Bandit Algorithm for Meal Recommendation}
Here we describe how we utilized contextual bandits and reinforcement learning to recommend meals to users. In our application of the boosted-bandit algorithm~\cite{kakadiya2021relational}, contextual bandits automatically extract and learn user preferences and provide highly personalized meal recommendations. Given a set of users with their dietary preferences($U$) as well as the set of all recipes in R3 format($R$), we create predicate logic pairs of the form \textit{preference(user\_5, negative\_nuts)} and \textit{item(food\_18, has\_nuts)}, indicating that the $user\_5$ has a negative preference for items with nuts and that food item 18 in $R$ has nuts. Following this, we also create negative and positive predicate logic pairs of the form \textit{recommendation(user\_18, food\_22)}, to signify if a recommendation is a positive or negative recommendation to the user based on their aligning preferences. We then split these predicate logic pairs into train and test sets and train the boosted-bandits ~\cite{kakadiya2021relational}. Then, we test the bandits on the test set, and for each \textit{recommendation(user, item)} pair, we receive a corresponding probability, which represents how likely the user is to prefer that item. We then construct a recommendation, $\mathcal{MP}$, by utilizing the users' preferred recommendation output (see \textbf{Problem Formulation} for details) and the highest probability items.

\newpage
\section{BEACON Implementation}

\subsection{System Architecture and Design}
This section describes the architectural design of the web application, which employs a hybrid approach combining React for the frontend and Django for the backend. This design aims to leverage the strengths of both frameworks to optimize system performance, user experience, and scalability. Additionally, the system incorporates a combination of relational and semi-structured data storage formats to efficiently manage diverse data types, including user profiles, recipes, and meal plans.

\begin{enumerate}
    \item \textbf{Frontend Architecture}: On the frontend, the system utilizes React to implement a dynamic Single Page Application (SPA) architecture. React's component-based design facilitates the development of a highly responsive user interface, enabling seamless navigation and interaction without requiring full-page reloads. This SPA-like behavior enhances the user experience, especially for features such as recipe navigation and interactive data visualizations. For complex data visualizations, the system integrates D3.js, a powerful JavaScript library that enables the rendering of sophisticated, data-driven graphics directly within the user's browser. By offloading the processing of visualizations to the client-side, this approach minimizes the load on the server, thereby improving scalability and responsiveness. Jest is employed for unit testing of React components, ensuring the reliability and maintainability of the frontend codebase. Dependency management for the frontend is handled through npm, which facilitates the installation and resolution of JavaScript package dependencies.

    \item \textbf{Backend Architecture}: On the backend, Django is employed to manage server-side operations, including user authentication, data processing, and the execution of machine learning algorithms for personalized meal plan generation. Django is chosen for its robustness, scalability, and comprehensive support for web application development. The backend will interact with AWS EC2 instances for scalable storage of user data, meal plans, and recipe information. The Django framework facilitates the creation of RESTful APIs through the Django REST Framework (DRF), enabling seamless communication between the frontend and backend. The use of DRF ensures that the system can efficiently handle requests from the frontend and manage data exchange in a structured manner.

    \item \textbf{Data Management}:
    \begin{itemize}
        \item \textbf{Users}: User data is stored in a relational database, with a schema designed to capture detailed attributes related to user profiles. Each user record includes essential fields such as name, email, username, and UUID (serving as the primary key). Additionally, the schema includes attributes for capturing user-specific health conditions, dietary restrictions, and food preferences, such as height, weight, isVegan, isDiabetic, as well as individual food preferences and allergies (e.g., likesDairy, allergicToGluten). This relational format ensures the integrity and consistency of user data.
        \item \textbf{Recipes}: Recipes are stored in a NoSQL database using a JSON-based schema (R3 format), designed to capture the complexity and variability of recipe data. The recipe schema includes fields such as recipe\_name, food\_role, and macronutrients. Recipes also include detailed ingredient lists, each of which specifies ingredient quantities, potential allergens, and associated image paths. Cooking instructions are stored in an array, allowing for the inclusion of multiple steps, each with relevant metadata (e.g., action name, output quality, background knowledge). This semi-structured format facilitates flexibility in the representation of diverse recipe data.
        \item \textbf{Meal Plans}: Meal plan data is also stored in a NoSQL database in JSON format. Each meal plan is organized by day, with each day's entries containing details about meals, including meal\_name, beverage, main\_course, side, and dessert. This format supports flexibility in meal organization and allows for easy modification and retrieval of meal plans for users.
    \end{itemize}

    \item \textbf{Scalability and Infrastructure}: To ensure the system is capable of scaling effectively, AWS EC2 instances are employed to host both the frontend and backend components of the application. AWS EC2 offers scalable cloud infrastructure, allowing the system to dynamically allocate resources based on varying levels of demand. This ensures that the web application can accommodate increasing user traffic without compromising performance. EC2's flexibility enables the system to scale horizontally, with additional instances provisioned to meet the needs of high-traffic periods. Additionally, the system benefits from the security and reliability features provided by AWS, including data redundancy, automatic load balancing, and secure data storage.

    The use of EC2 allows for seamless scaling of both compute and storage resources, which is critical as the application evolves and the user base grows. As a result, the architecture is designed to handle future demands and provide the necessary infrastructure to support a high-availability, high-performance application.

\end{enumerate}

\subsection{System Figures}

\begin{figure*}
    \centering
    \includegraphics[width=1\linewidth]{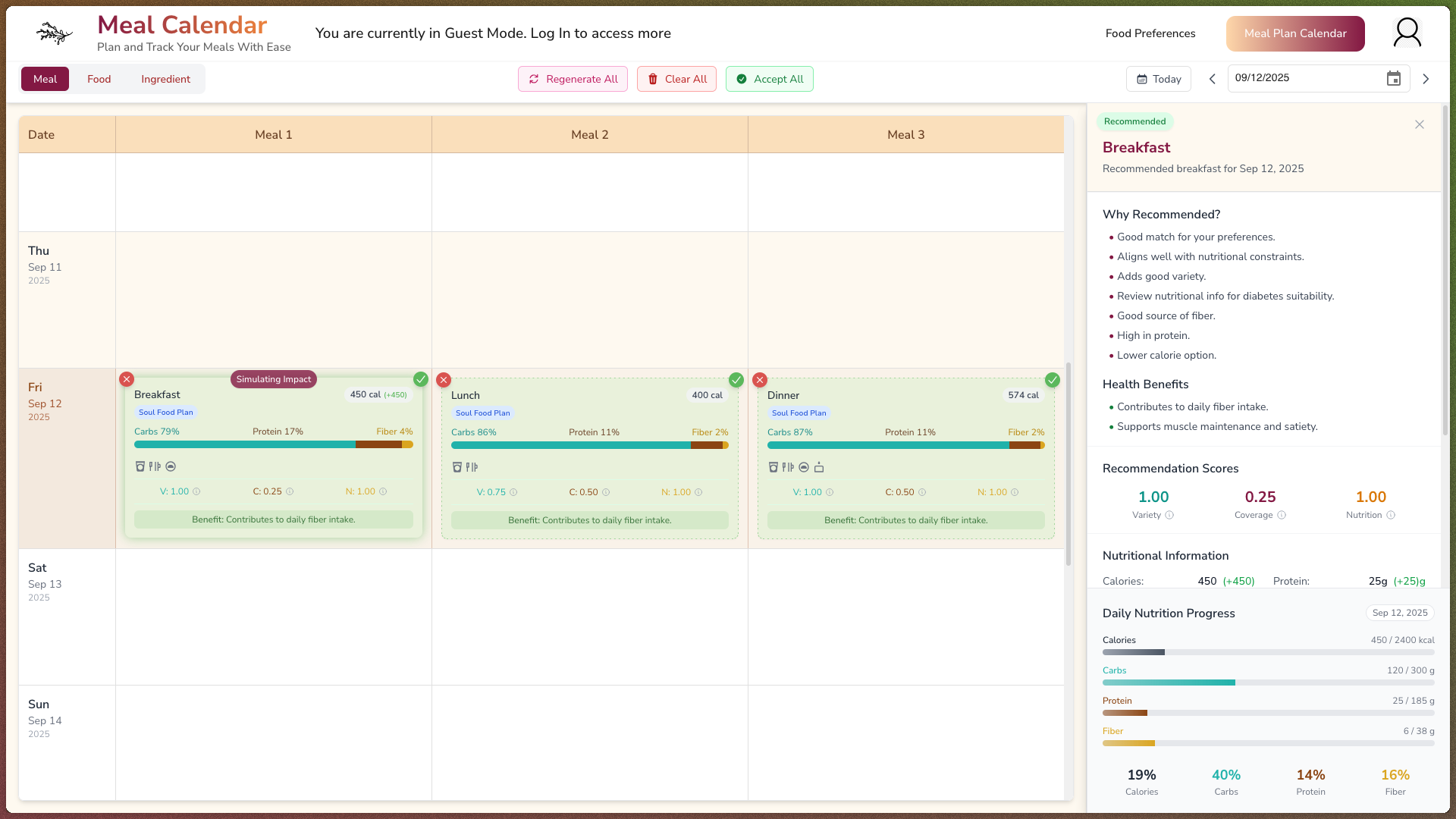}
    \caption{Calendar-based meal visualization screen on the \textbf{Meal View}.}
    \label{fig:meal}
\end{figure*}

\begin{figure*}
    \centering
    \includegraphics[width=1\linewidth]{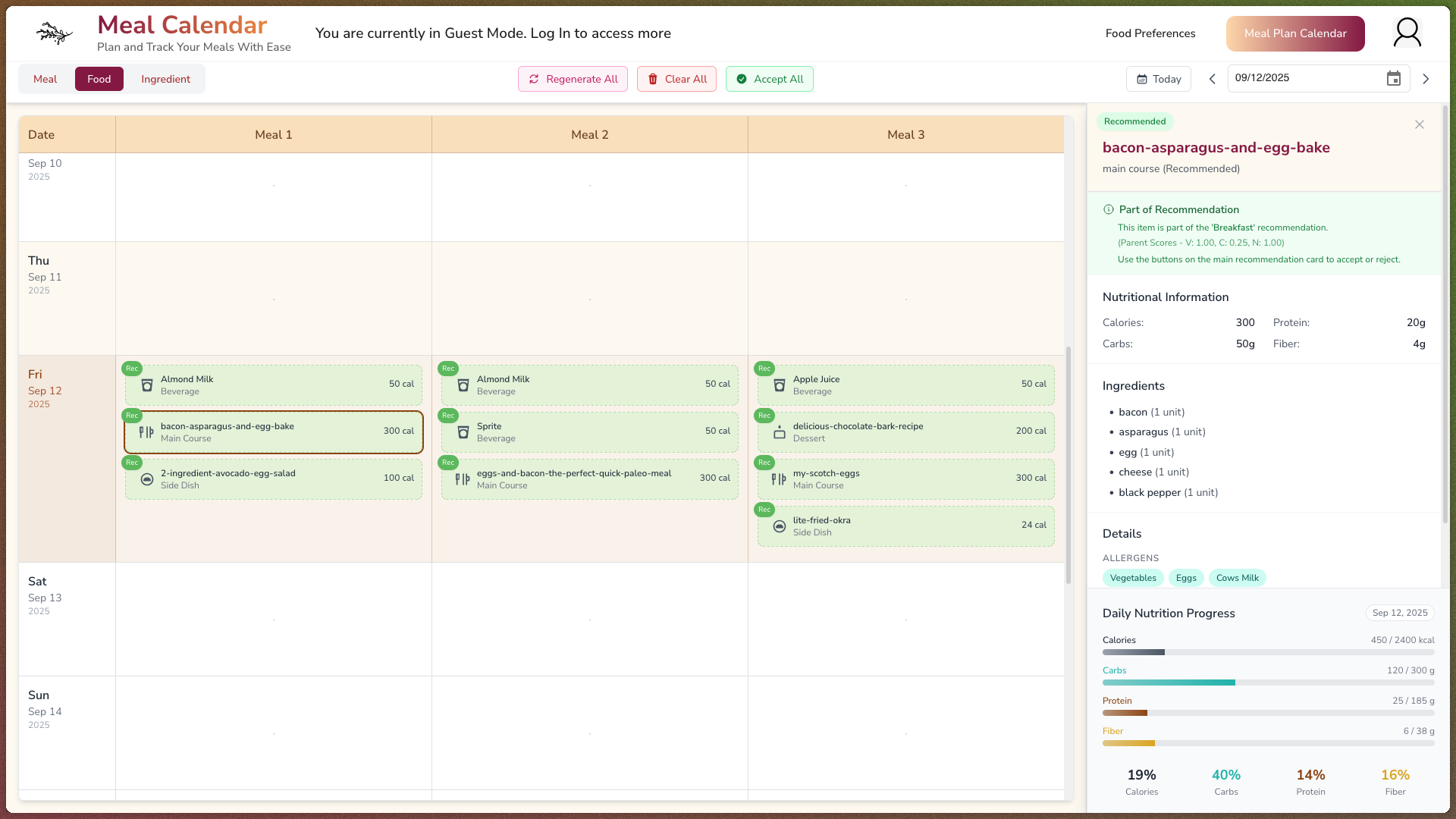}
    \caption{Calendar-based meal visualization screen on the \textbf{Food Item View}.}
    \label{fig:food}
\end{figure*}

\begin{figure*}
    \centering
    \includegraphics[width=1\linewidth]{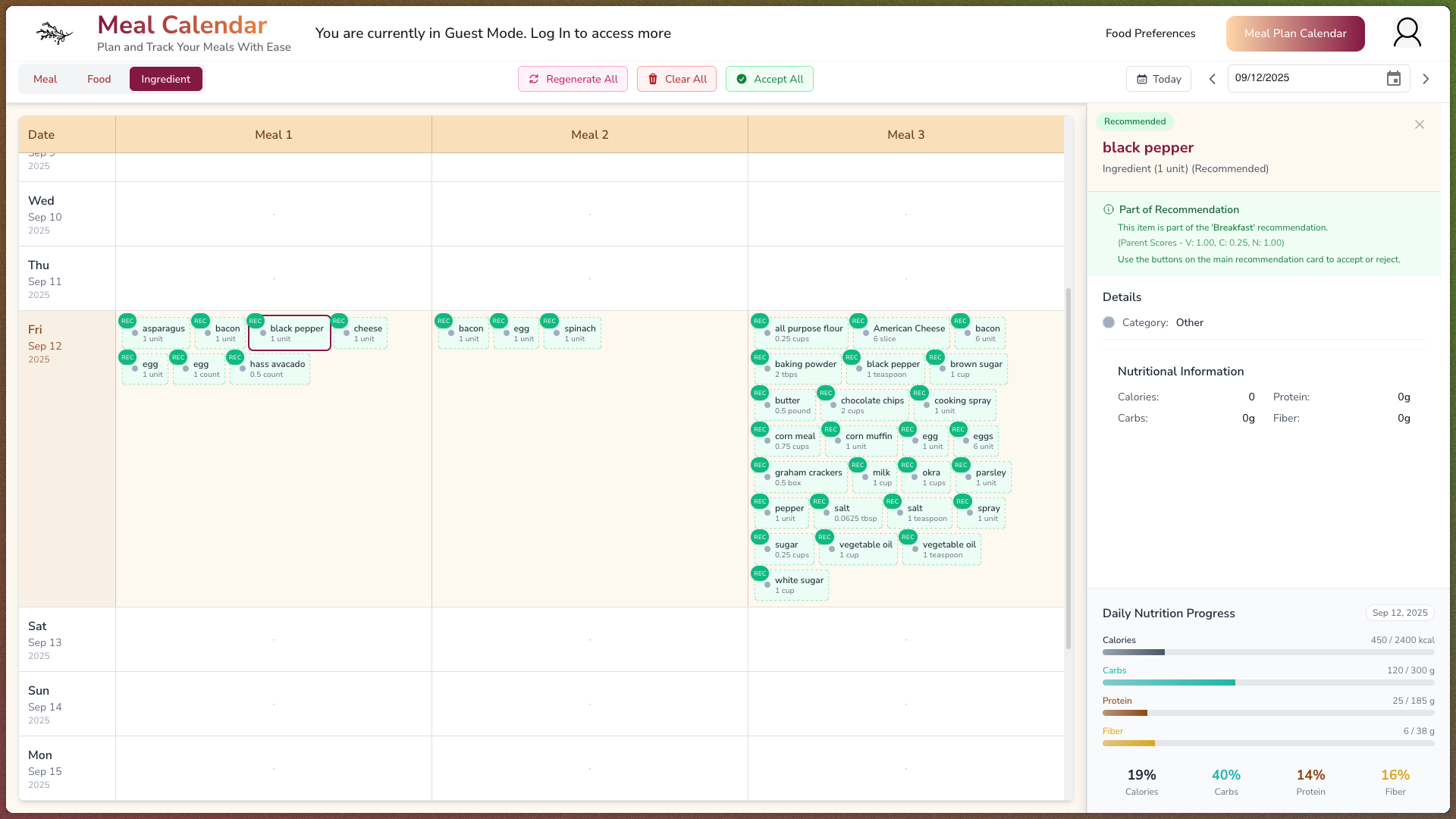}
    \caption{Calendar-based meal visualization screen on the \textbf{Ingredient View}.}
    \label{fig:ingredient}
\end{figure*}

Figures~\ref{fig:meal}, \ref{fig:food}, and \ref{fig:ingredient} show the meal, food, and ingredient views of BEACON's interactive meal calendar, respectively.

\end{document}